\documentclass[10pt,twocolumn,letterpaper]{article}

\usepackage{wacv}
\usepackage{times}
\usepackage{epsfig}
\usepackage{graphicx}
\usepackage{multirow}
\usepackage{subcaption}
\usepackage{amsmath}
\usepackage{amssymb}
\usepackage{listings}

\usepackage{xcolor}
\usepackage[normalem]{ulem}

\wacvfinalcopy 

\def\wacvPaperID{362} 

\ifwacvfinal\fi
\newcommand{\lib}{\textit{Kornia}}

\begin{document}

\title{\lib: an Open Source Differentiable Computer Vision Library for PyTorch}

\author{Edgar Riba \\
Computer Vision Center\\
OSVF-OpenCV.org\\
{\tt\small edgar.riba@gmail.com}
\and
Dmytro Mishkin \\
VRG, Faculty of Electrical Engineering\\
Czech Technical University in Prague\\
{\tt\small ducha.aiki@gmail.com}
\and
Daniel Ponsa \\
Computer Vision Center\\
{\tt\small daniel@cvc.uab.es}
\and
Ethan Rublee \\
Arraiy, Inc.\\
{\tt\small ethan.rublee@gmail.com}
\and
Gary Bradski \\
OSVF-OpenCV.org\\
{\tt\small garybradski@gmail.com}
}

\maketitle
\ifwacvfinal\thispagestyle{empty}\fi

\begin{abstract}
   This work presents \lib{} -- an open source computer vision library which consists of a set of differentiable routines and modules to solve generic computer vision problems. The package uses \textit{PyTorch} as its main backend both for efficiency and to take advantage of the reverse-mode auto-differentiation to define and compute the gradient of complex functions. Inspired by \textit{OpenCV}, \lib{} is composed of a set of modules containing operators that can be inserted inside neural networks to train models to perform image transformations, camera calibration, epipolar geometry, and low level image processing techniques, such as filtering and edge detection that operate directly on high dimensional tensor representations. Examples of classical vision problems implemented using our framework are provided including a benchmark comparing to existing vision libraries.
\end{abstract}

\vspace{-1em}
\section{Introduction}

\begin{figure}[t]
    \setlength\tabcolsep{2.5pt}
    \begin{center}
        \begin{tabular}{c c c}
        \textbf{Color} & \textbf{Filtering} & \textbf{Geometry}\\
        \includegraphics[width=2.5cm]{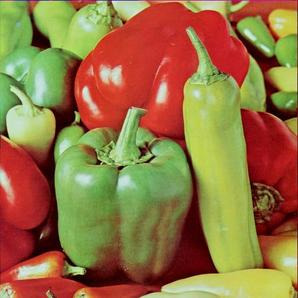} &
        \includegraphics[width=2.5cm]{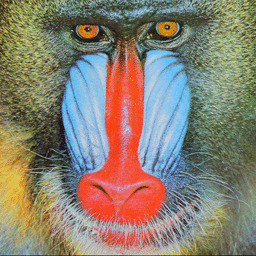} &
        \includegraphics[width=2.5cm]{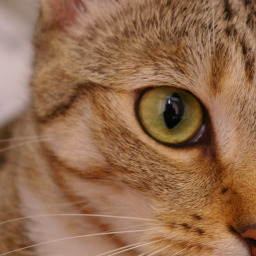} \\
        \includegraphics[width=2.5cm]{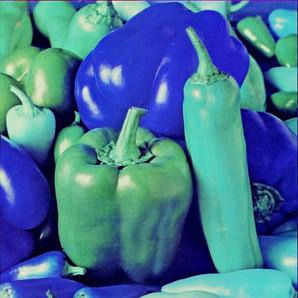} &
        \includegraphics[width=2.5cm]{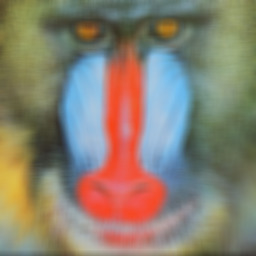} &
        \includegraphics[width=2.5cm]{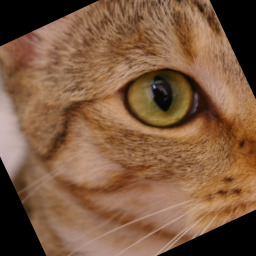} \\
        \includegraphics[width=2.5cm]{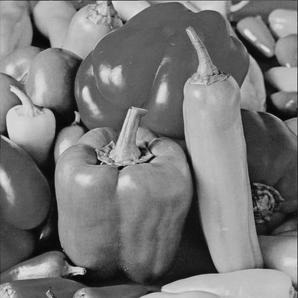} &
        \includegraphics[width=2.5cm]{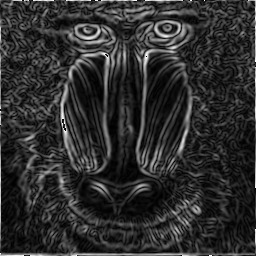} &
        \includegraphics[width=2.5cm]{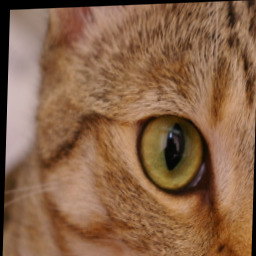} \\
        \end{tabular}
    \end{center}
    \vspace{-1.5em}
    \caption{The library implements routines for low level image processing tasks using native \textit{PyTorch} operators and their custom optimization. The purpose of the library is to be used for large-scale vision projects, data augmentation, or for creating computer vision layers inside of neural network layers that allow for backprogating error through them. The above results are obtained from a given batch of images using data parallelism in the GPU.}
    \label{fig:imgproc}
\end{figure}
Computer vision has driven a lot of advances in modern society for different industries such as self driving cars, industrial robotics, visual effects, image search, etc resulting in a wide field of applications. One of the key components of this achievement has been due to the open-source software and the community that helped to make all this possible by providing open-source implementations of the main computer vision algorithms.

There exist several open-source libraries widely used by the computer vision community designed and optimized to process images using Central Processing Units (CPUs). However, many of the best performing computer vision algorithms are now based on deep learning, processing images in parallel using Graphical Processing Units (GPUs). Within that context, a framework that is gaining popularity is Pytorch \cite{pytorch} due to its reverse-mode automatic differentiation mechanism, dynamic computation graph, distributed learning and eager/script execution modes. PyTorch and its ecosystem provide a few packages to work with images such as it's most popular toolkit, \textit{torchvision}, which is mainly designed to perform data augmentation, read popular datasets and implementations of state-of-the-art models for tasks such as detection, segmentation, image generation, and landmark detection. Yet, it lacks implementations for standard vision algorithms that can be run directly on GPUs using their native tensor data structures.
\begin{table*}[t]
\begin{tabular}{l*{6}{c}r}
                   & CPU & GPU & Batch Processing & Differentiable & Distributed & Multi-dimensional array \\
\hline
torchvision\cite{pytorch} & \checkmark & $\times$ & $\times$ & $\times$ & $\times$ & $\times$ \\
scikit-image~\cite{scikit-image} & \checkmark & $\times$ & $\times$ & $\times$ & $\times$ & $\times$ \\
opencv~\cite{opencv}       & \checkmark & \checkmark & $\times$ & $\times$ & $\times$ & $\times$ \\
tensorflow.image~\cite{tensorflow2015-whitepaper}   & \checkmark & \checkmark & \checkmark & \checkmark & \checkmark & \checkmark \\
\lib{}   & \checkmark & \checkmark & \checkmark & \checkmark & \checkmark & \checkmark \\ \\
\end{tabular}
    \vspace{-1.5em}
\caption{\label{tab:table-cv-frameworks} Comparison of  different computer vision libraries by their main features. \lib{} and tensorflow.image are the only frameworks that mostly implement their functionalities in the GPU, using batched data, differentiable and have the ability to be distributed.}
    \vspace{-1.5em}
\end{table*}

This paper introduces \lib, an open source computer vision library built on top of PyTorch that will help students, researchers, companies and entrepreneurs to implement computer vision applications oriented towards deep learning. Our library, in contrast to traditional CPU-based vision frameworks, provides standard image processing functions implemented on GPUs that can also be embedded inside deep networks.

\lib{} is designed to fill the gap between PyTorch and computer vision communities and it is based on some of the pre-existing open source solutions for computer vision (PIL, skimage, torchvision, tf.image), but with a strong inspiration on OpenCV~\cite{opencv}. \lib{} combines the simplicity of both frameworks in order to leverage differentiable programming for computer vision taking properties from PyTorch such as differentiability, GPU acceleration, or  distributed data-flows.

In addition to introducing \lib, this paper contributes some demos showing how \lib{} eases the implementation of several computer vision tasks like image registration, depth estimation or local features detection which are common in many computer vision systems.

The rest of the paper is organized as follows: we review the state of the art in terms of open source software for computer vision and machine learning in Section \ref{section:related_work}; Section \ref{section:kornia} describes the design principles of the proposed library and all its components, and Section~\ref{section:use_cases} introduces use cases that can be implemented using the library's main features.

\section{Related work}
\label{section:related_work}
We present in this section a review of the state of the art for computer vision software. Related works will be divided in two main categories: traditional computer vision and deep learning oriented computer vision frameworks. The first with a focus on the very first libraries that implement mostly algorithms optimized for the CPU, and the second targeting solutions for GPU.

\subsection{Traditional computer vision libraries}
\label{section:related_work:traditional_vision}
Nowadays there are many different frameworks that implement computer vision algorithms. However, during the early days of computer vision, it was difficult to find any centralized software with image processing algorithms. All the existing software for computer vision was mostly developed within universities or at small teams in companies, not shipped in any form and neither released to the public domain.

It was not until Intel released the first version of the Open Source Computer Vision Library (OpenCV). OpenCV \cite{opencv} which initially implemented computer vision algorithms for real-time ray tracing, visual interfaces and 3D display walls. All the algorithms were made available with a permissive library not only research, but also production. OpenCV changed the paradigm within the computer vision community given the fact that most of the state of art algorithms in computer vision were now put in an common framework written very efficient in C, becoming in that way a reference within the community.

The computer vision community shifted to improving or besting existing algorithms and started sharing their code with the community. This resulted in new code optimized mostly for CPU. Vedaldi et al. introduced \textit{VLFeat}~\cite{vedaldi08vlfeat}, an open source library that implements popular computer vision algorithms specializing in image understanding and local features extraction and matching. VLFeat was written in C for efficiency and compatibility, with interfaces in MATLAB. For ease of use, it supported Windows, Mac OS X, and Linux, and has been a reference e.g for efficient implementations of algorithms such as Fisher Vector~\cite{Sanchez2013}, VLAD~\cite{VLAD2010}, SIFT~\cite{Lowe2004}, and MSER~\cite{Matas2002}.

MathWorks released a proprietary Computer Vision Toolbox inside one of its famous product MATLAB~\cite{MATLAB:2010} that covered many of the main computer vision, 3D vision, and video processing algorithms which has been used by many computer vision students and researchers becoming quite standard within the researcher community. The computer vision community have been using to MATLAB for some decades, and many still use it.

Existing frameworks like Scikit-learn \cite{scikit-learn} partially implement machine learning algorithms used by the computer vision community for classification, regression and clustering including support vector machines, random forests, gradient boosting and k-means. Similar project as Scikit-image~\cite{scikit-image} implement open source collections of algorithms for image processing.

\subsection{Deep learning and computer vision}
\label{section:related_work:deep_learning}
Computer vision frameworks have been optimized for CPU to fulfill realtime applications, but the recent success of deep learning in the field object classification changed the way of addressing many traditional computer vision tasks. A. Krizhevsky et al~\cite{AlexNet2012} took the old ideas from Yann LeCun's Convolutional Neural Networks (CNNs)~\cite{MNIST1998} paper with an architecture similar to LeNEt-5 and achieved the best results by far in the ILSVRC~\cite{ILSVRC15} image classification task. This was a breakthrough moment for the computer vision community, and changed the way computer vision was understood. In terms of software, new frameworks such \textit{Caffe} \cite{caffe}, Torch \cite{torch7}, MXNet~\cite{journals/corr/ChenLLLWWXXZZ15}, Chainer~\cite{Tokui:2019:CDL:3292500.3330756}, Theano~\cite{Bergstra10theano}, MatConvNet~\cite{Vedaldi15}, PyTorch~\cite{pytorch}, and Tensorflow \cite{tensorflow2015-whitepaper} appeared on the scene implementing many old ideas in the GPU using parallel programming~\cite{CookCUDA} as an approach to handle the need for large amounts of data processing in order to train deep learning models.

With the rise of deep learning, most of the standard computer vision frameworks have moved to being used more for certain geometric vision functions, data pre-processing, data augmentation on the CPU in order to be transferred later to the GPU as well as post processing to refine results. Examples of libraries that are currently used to perform pre and post-processing on the CPU within the deep learning frameworks are OpenCV or PIL.

Given that most of the deep learning frameworks still use standard vision libraries to perform the pre and post processing on CPU and similar to Tensorflow.image, {as Table~\ref{tab:table-cv-frameworks} shows, we fill the gap} within the PyTorch ecosystem introducing a computer vision library that implements standard vision algorithms taking advantage of the different properties that modern frameworks for deep learning like PyTorch can provide: 1) \textit{differentiability} for commodity avoiding to write derivative functions for complex loss functions; 2) \textit{transparency} to perform parallel or serial computing either in CPU or GPU devices using batches in a common API; 3) \textit{distributed} for computing large-scale applications; 4) code ready for \textit{production}. For this reason, we present \lib, a modern computer vision framework oriented for deep learning.

\begin{figure*}
\centering
   \includegraphics[width=0.4\linewidth]{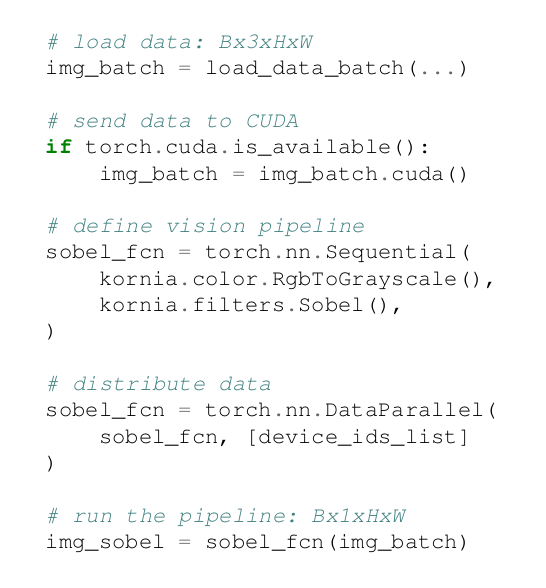}
\includegraphics[width=0.45\linewidth]{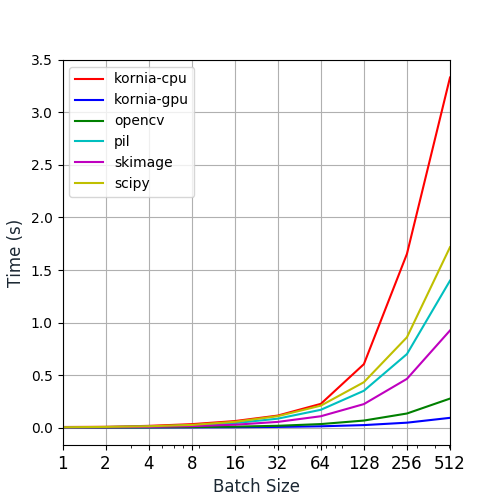}
\caption{\textbf{Left:} Python script showing our image processing API. Notice that the API is transparent to the device, and can be easily combined with other PyTorch components. \textbf{Right:} Results of the benchmark comparing \lib{} to other state-of-the-art vision libraries. We measure the elapsed time for computing Sobel edges (lower is better).}
\label{fig:imgproc:listing_benchmark}
\end{figure*}
\section{\lib: Computer Vision for PyTorch.}
\label{section:kornia}
\lib\footnote{\url{https://kornia.org}} can be defined as a computer vision library for PyTorch, inspired by OpenCV and with strong GPU support. \lib{} allows users to write code as they were using plain PyTorch providing high level interfaces to vision algorithms computed directly on tensors. In addition, some of the main PyTorch features are inherited by \lib{} such as a high performance environment with easy access to automatic differentiation, executing models on different devices (CPU and GPU), parallel programming by default, communication primitives for multiprocess parallelism across several computation nodes and code ready for production. In the following, we remark these properties.

\textbf{Differentiable}. An image processing algorithm that can be defined as a Direct Acyclic Graph (DAG) structure can, thanks to the reverse-mode \cite{Speelpenning1980CFP} auto-differentiation \cite{Griewank:2008:EDP:1455489}, compute gradients via backpropagation \cite{kelley1960gradient}. In practice, this means that such computer vision functions are operators that can be placed as layers within the neural networks for training via backpropagating through them.

\textbf{Transparent API}. A key component in the library design is its easy way to seamlessly add hardware acceleration to your program with a minimum of effort. The library API is agnostic to the input source device, meaning that the algorithms can either be run in CPU or GPU.

\textbf{Parallel programming}. Batch processing is another important feature that enables running vision operators using data parallelism by default. The assumption for the operators is to receive as input batches of N-channel image tensors, contrary to standard vision libraries with single 1-3 channel images. Hence, for Kornia working with multispectral or hyperspectral images would be direct.

\textbf{Distributed}. Support for communication primitives for multi-process parallelism across several computation nodes running on one or more machines. The library design allows users to run their applications in different distributed systems, or even able to process large vision pipelines in an efficient way.

\textbf{Production}. Since its latest versions, PyTorch is able to serialize and optimize models for production purposes. Based on its just-in-time (JIT) compiler, PyTorch traces the models creating \textit{TorchScript} programs at runtime in order to be run in a standalone C++ program using kernel fusion to do faster inference making out library a perfect fit also for built-in vision products.

\subsection{Library structure}
\label{section:kornia:library_structure}

Similar to other frameworks, the library is composed of several submodules grouped by generic computer vision topics:

\textbf{|kornia.color:|} provides operators for color space conversions. The functionality found in this module covers conversions such as Grayscale, RGB, BGR, HSV, YCbCr. In addition, operators to adjust color properties such as brightness, contrast hue or saturation are also provided.

\textbf{|kornia.features:|} provides operators to detect local features, compute descriptors, and perform feature matching. The module provides differentiable versions of the Harris corner detector\cite{Harris88}, Hessian detector~\cite{Hessian78}, their scale and affine covariant versions~\cite{Mikolajczyk2004}, DoG~\cite{Lowe2004}, patch dominant gradient orientation~\cite{Lowe2004} and the SIFT descriptor~\cite{Lowe2004}. |kornia.features| provides a high level API to perform detections in scale-space, where classical hard non-maxima suppression is replaced with its soft version similar to the recently proposed Multiscale Index Proposal layer (M-SIP)~\cite{KeyNet2019}. One can seamlessly replace any or all modules with deep learned counterparts. A set of operators for work with local features geometry is also provided.

\textbf{|kornia.filters:|} provides operators to perform linear or non-linear filtering operations on tensor images. Functions to convolve tensors with kernels, for computing first and second order image derivatives, or high level differentiable implementations for blurring algorithms such as Gaussian and Box blurs, Laplace, and Sobel\cite{kanopoulos1988design} edges detector.

\textbf{|kornia.geometry:|} module devoted to perform 2D and 3D geometry structured as follows:
\begin{itemize}
\setlength\itemsep{0em}
    \item \textbf{|transforms:|} A set of operators to perform geometrical image transformations such rotation, translation, scaling, shearing, and primitives for more complex affine and homography based transforms.
    \item \textbf{|camera:|} A set of routines specific to different types of camera representation such as Pinhole or Orthographic models containing functionality such as projecting and unprojecting points from the camera to the world frame.
    \item \textbf{|conversions:|} A set of routines to perform conversions between angle representation such as radians to degrees, coordinates normalization, and homogeneous to euclidean. Moreover, we include advanced conversions for 3D geometry representations such as Quaternion, Axis-Angle, RotationMatrix, or Rodrigues formula.
    \item \textbf{|linalg:|} A set of routines to perform general rigid-body homogeneous transformations. We include implementations to transform points between frames and for homogeneous transformations, manipulation such as composition, inverse and to compute relative poses.
    \item \textbf{|warp:|} A set of primitives to sample image tensors from a reference to a non-reference frame by its related homography or the depth.
\end{itemize}

\textbf{|kornia.losses:|} A stack of loss functions to be used to solve specific vision tasks such as semantic segmentation, and image reconstruction such as the Structural Similar Index Loss (SSIM)~\cite{Wang:2004:IQA:2319031.2320551}.

\textbf{|kornia.contrib:|} A set of experimental operators and user contributions containing routines for splitting tensors in blocks, or to perform subpixel accuracy like the softargmax2d operator.
\begin{figure*}
    \setlength\tabcolsep{2.5pt}
    \begin{center}
        \begin{tabular}{c c c c c c}
        \textbf{Level 1} & \textbf{Level 2} & \textbf{Level 3} & \textbf{Level 4} & \textbf{Level 5} & \textbf{Level 6} \\
        \includegraphics[width=2.5cm]{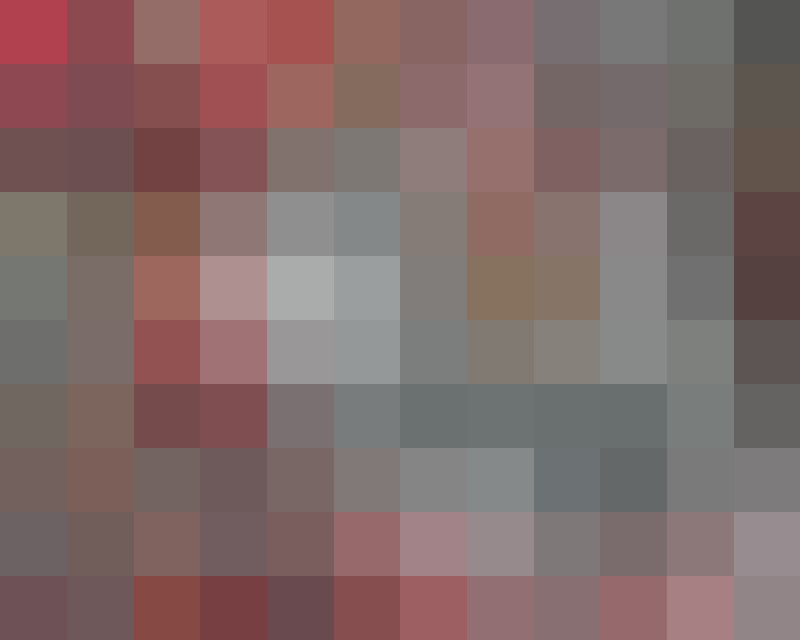} &
        \includegraphics[width=2.5cm]{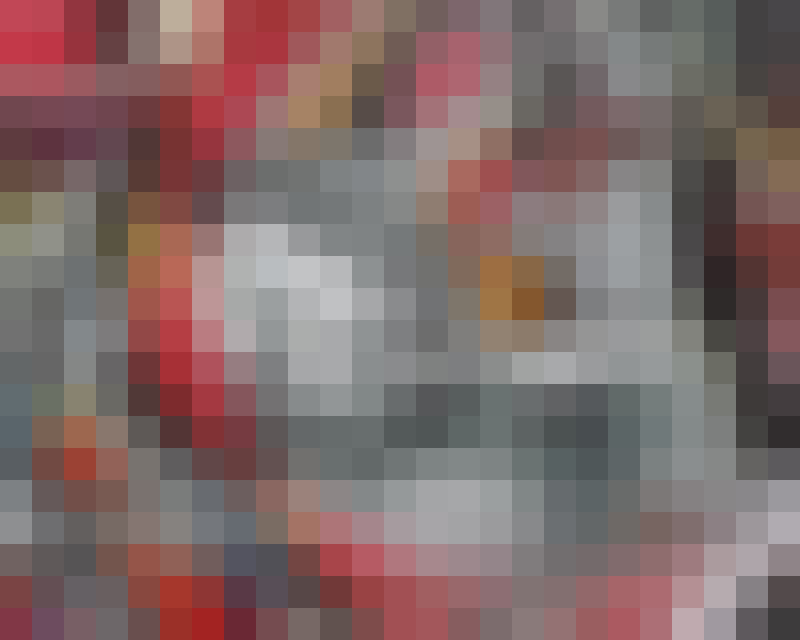} &
        \includegraphics[width=2.5cm]{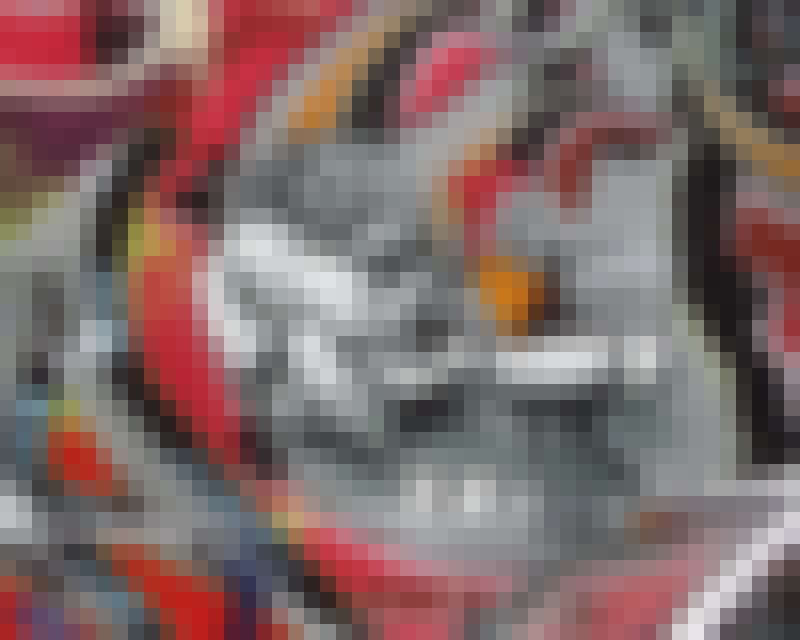} &
        \includegraphics[width=2.5cm]{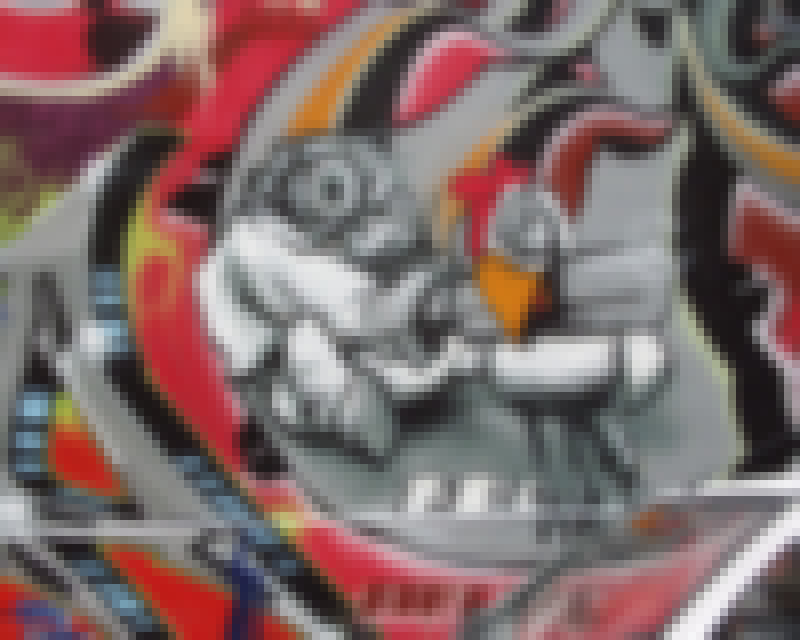} &
        \includegraphics[width=2.5cm]{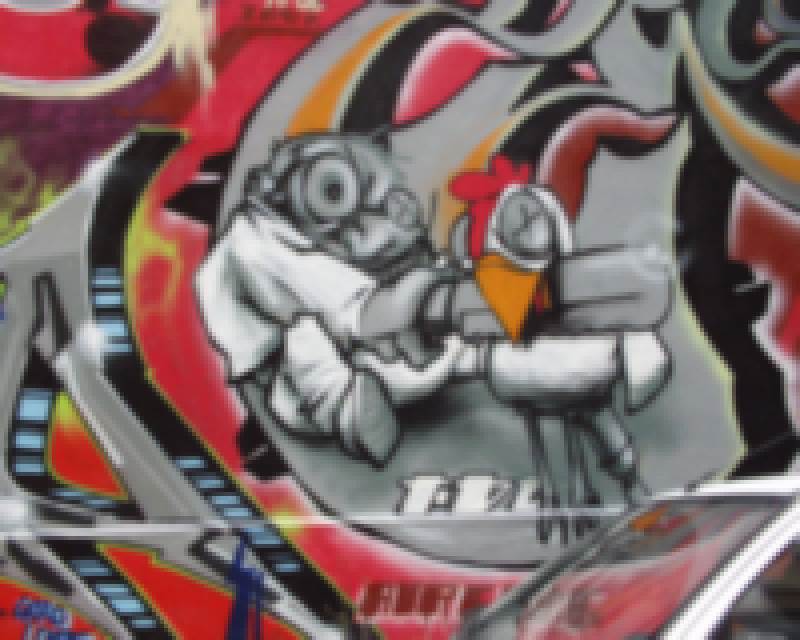} &
        \includegraphics[width=2.5cm]{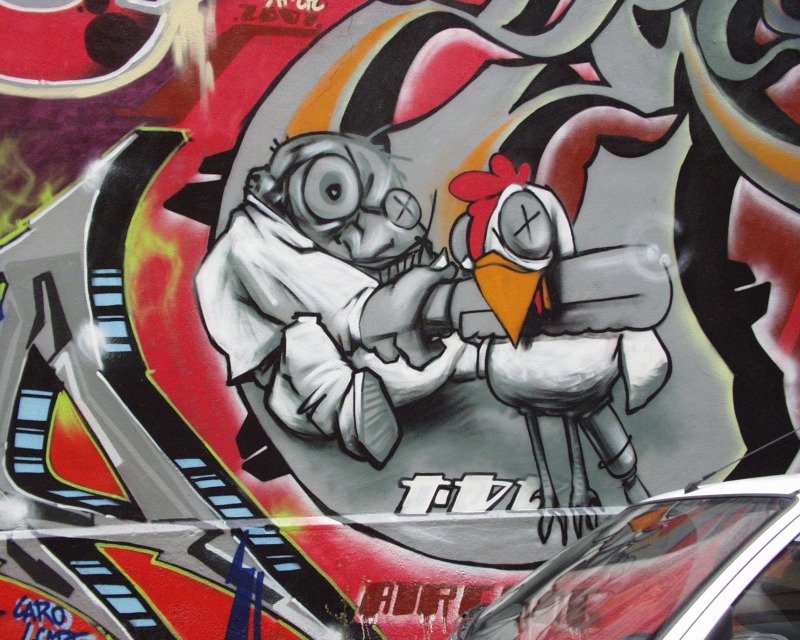} \\
        \includegraphics[width=2.5cm]{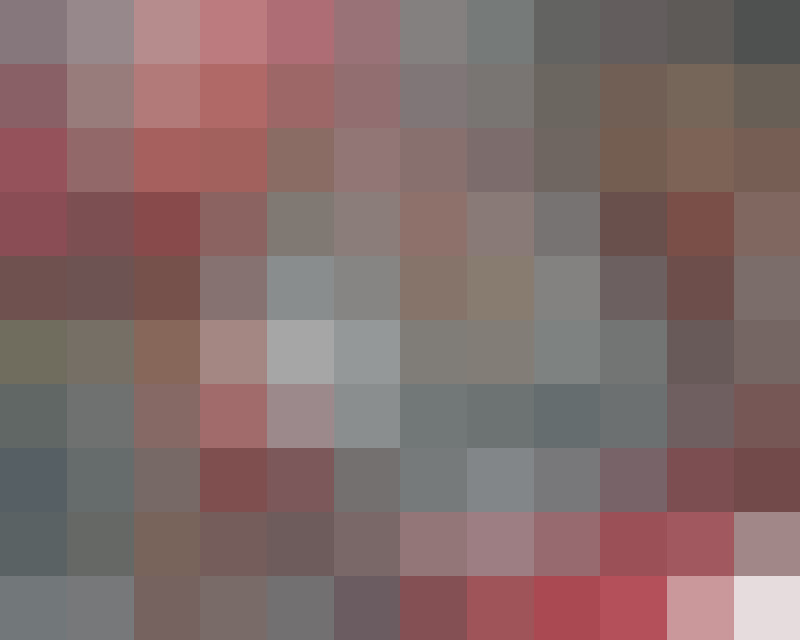} &
        \includegraphics[width=2.5cm]{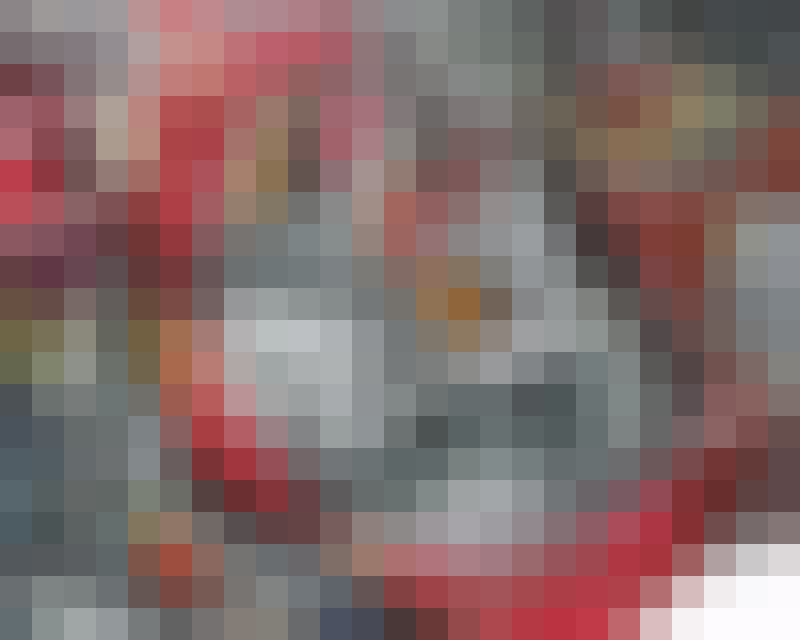} &
        \includegraphics[width=2.5cm]{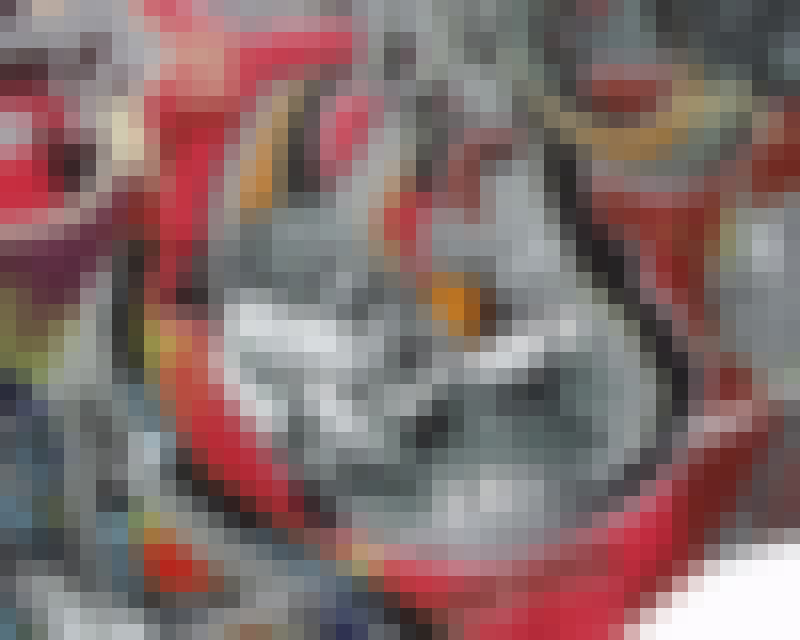} &
        \includegraphics[width=2.5cm]{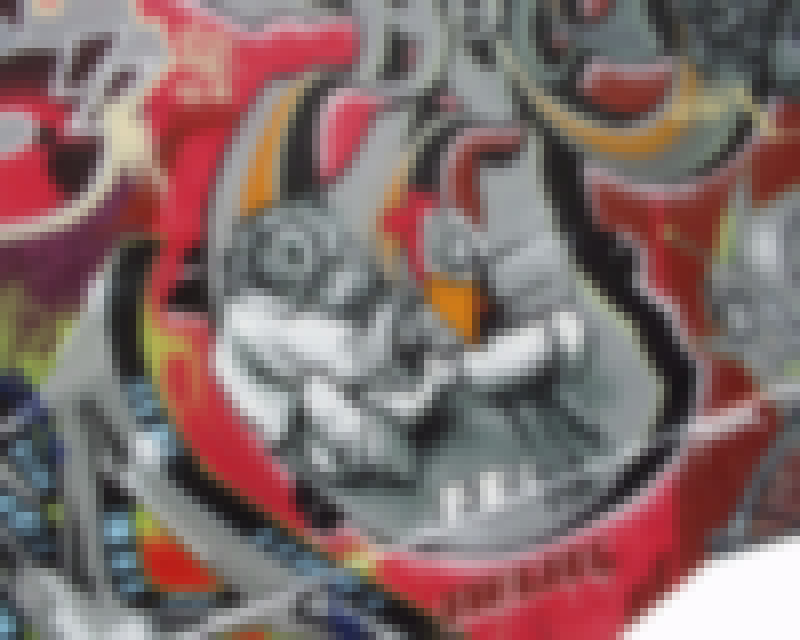} &
        \includegraphics[width=2.5cm]{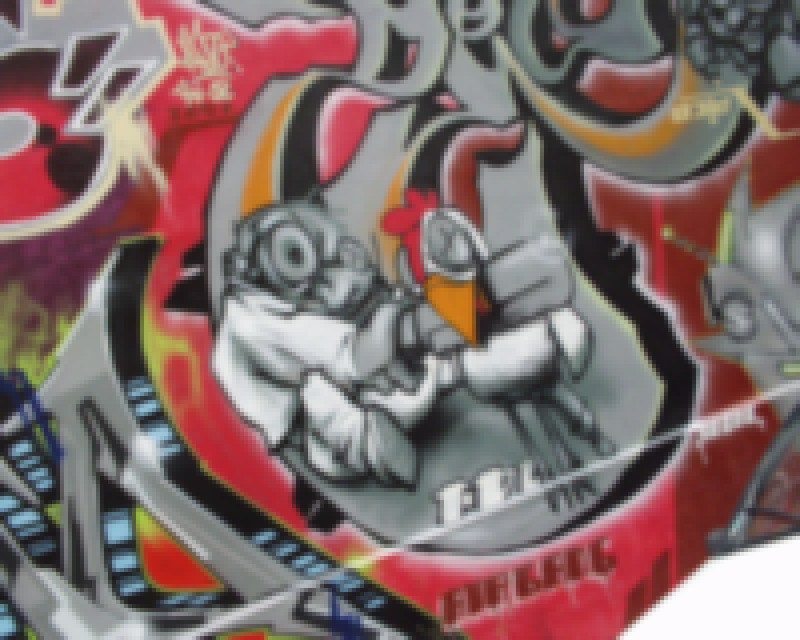} &
        \includegraphics[width=2.5cm]{data/homography_img_dst_7.jpg} \\
        \includegraphics[width=2.5cm]{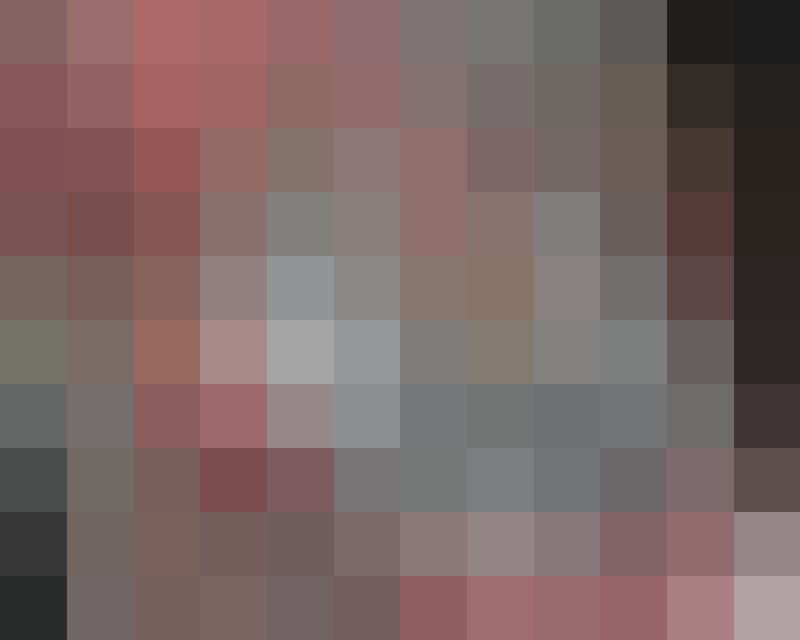} &
        \includegraphics[width=2.5cm]{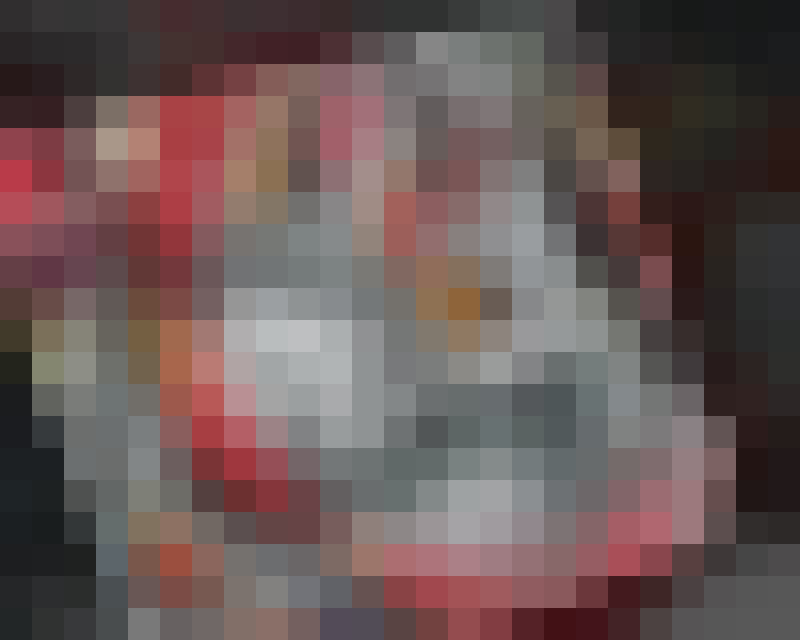} &
        \includegraphics[width=2.5cm]{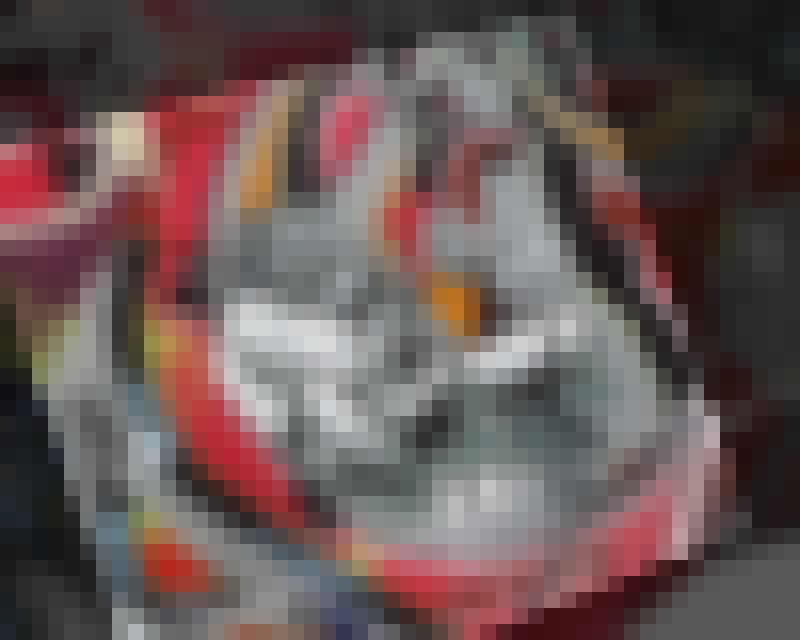} &
        \includegraphics[width=2.5cm]{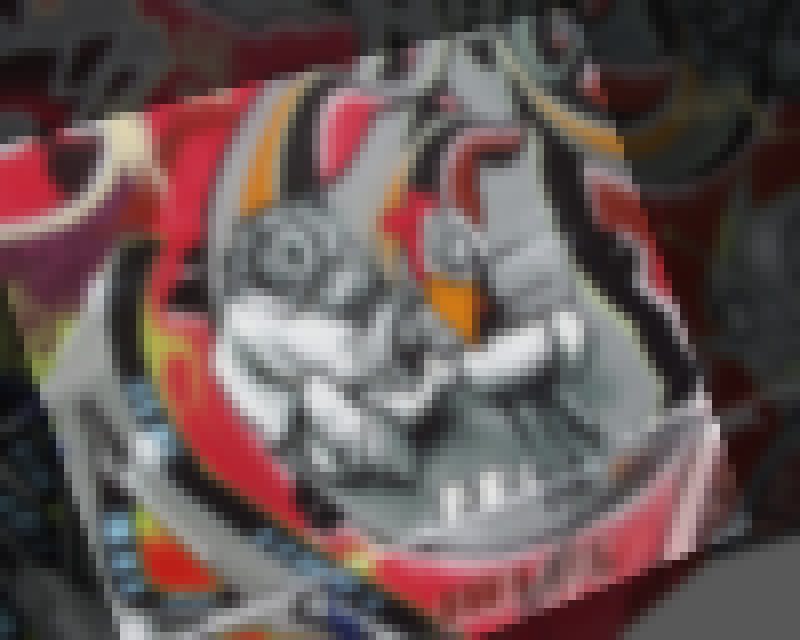} &
        \includegraphics[width=2.5cm]{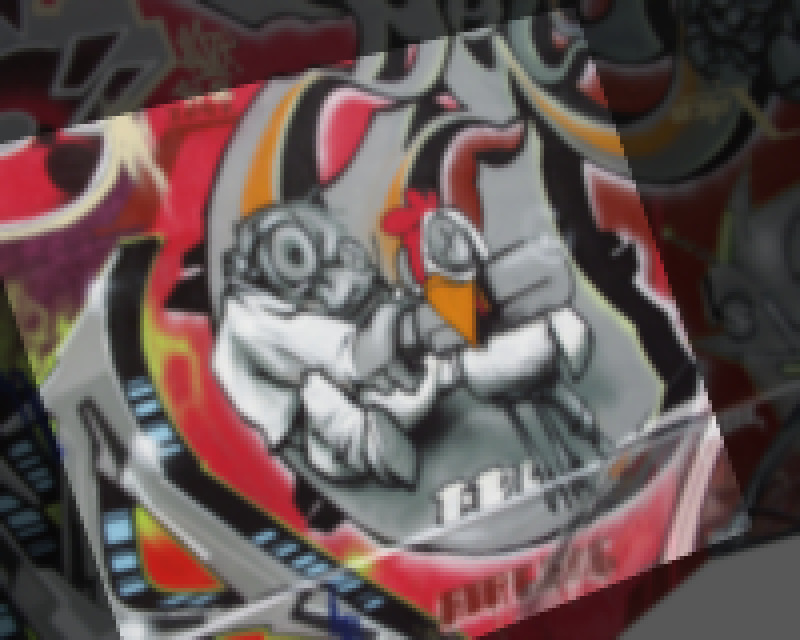} &
        \includegraphics[width=2.5cm]{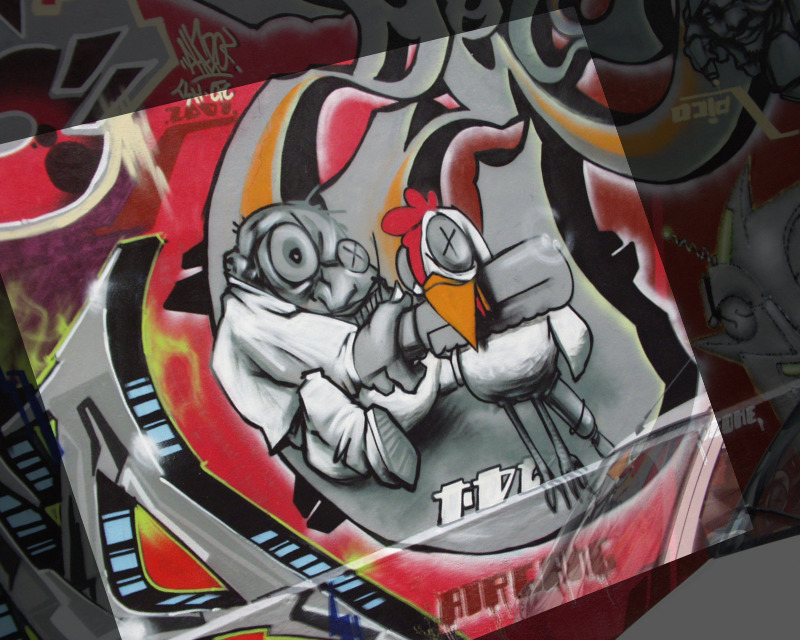} \\
        \includegraphics[width=2.5cm]{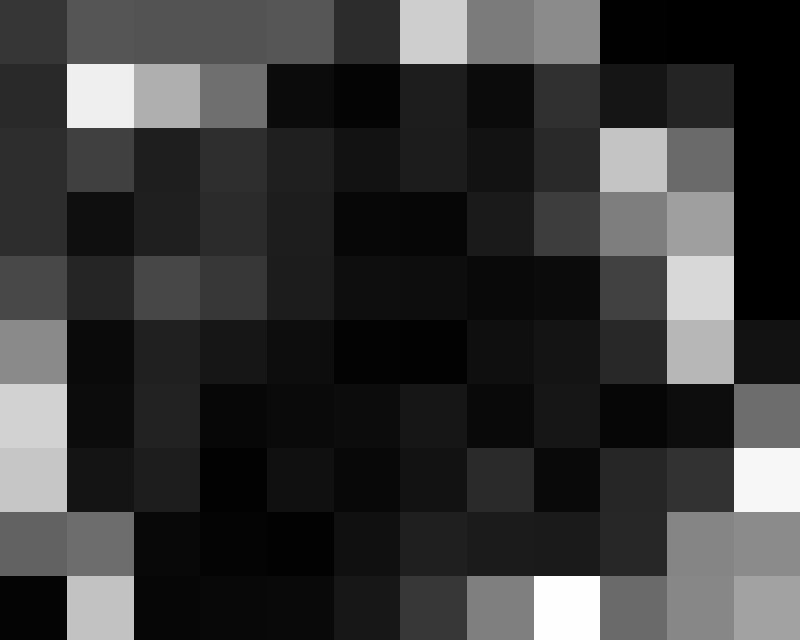} &
        \includegraphics[width=2.5cm]{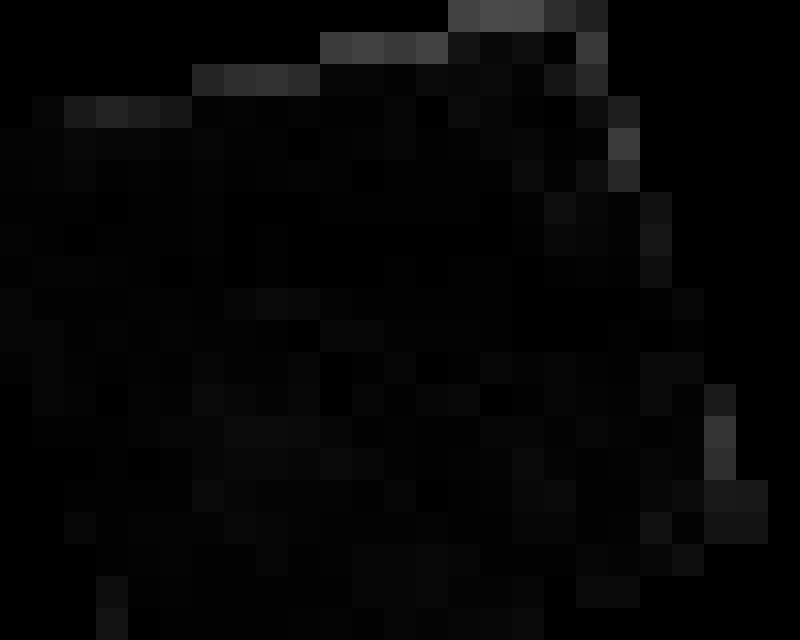} &
        \includegraphics[width=2.5cm]{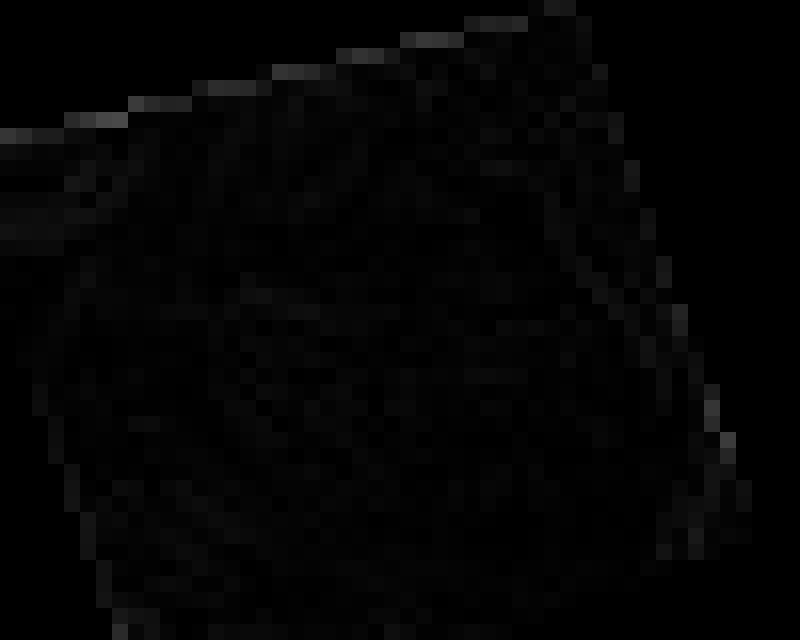} &
        \includegraphics[width=2.5cm]{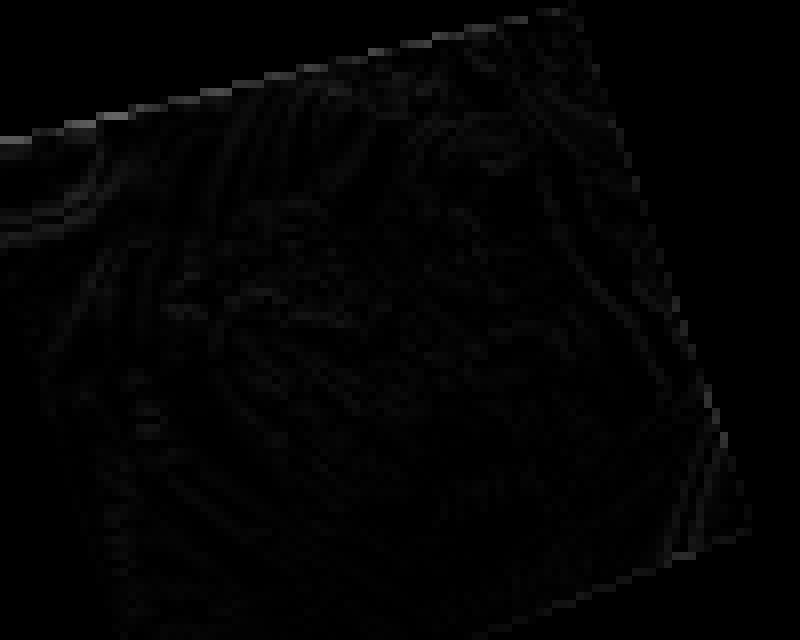} &
        \includegraphics[width=2.5cm]{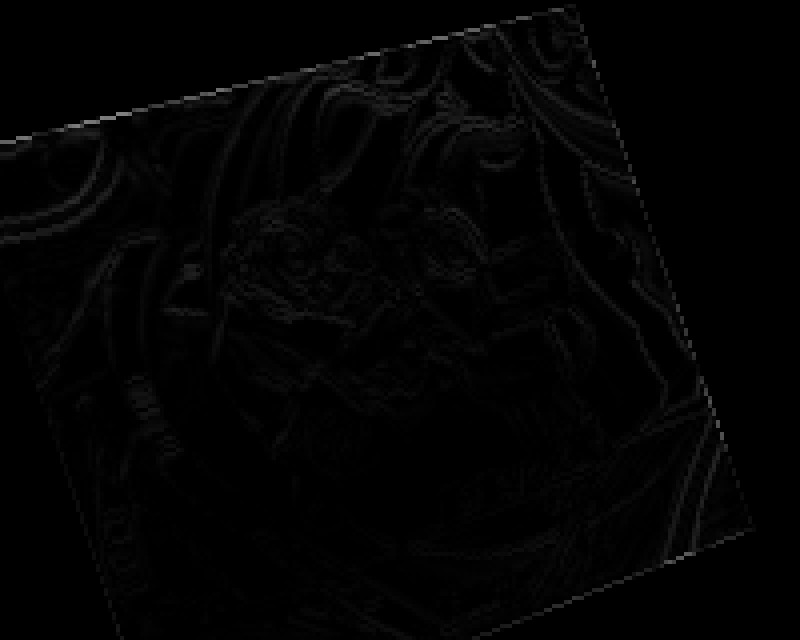} &
        \includegraphics[width=2.5cm]{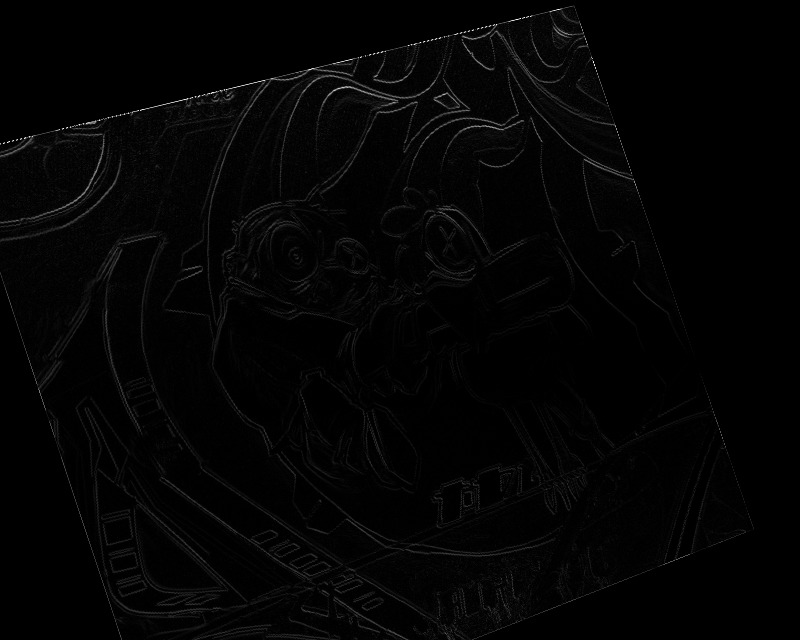} \\
        \end{tabular}
    \end{center}
    \caption{Results of the image registration by gradient descent. Each of the columns represent a different level of the image pyramid used to optimize the loss function. \textit{Row 1:} the original source image; \textit{Row 2:} the original destination image; \textit{Row 3:} the source image warped to destination at the end of the optimization loop at that specific scale level.\textit{ Row 4:} the photometric error between the warped image using the estimated homography and the warped image using the ground truth homography. The algorithm starts to converge in the lower scales refining the solution as it goes to the upper levels of the pyramid.}
    \label{fig:homography_pyramid}
\end{figure*}
\section{Use cases}
\label{section:use_cases}
This section presents practical examples of the library use for well known classical vision problems demonstrating its easiness for computing the derivatives of complex loss functions and releasing the user of that part. We first show quantitative and qualitative results on experiments comparing our image processing API compared to existing image processing libraries. Next, an example of image registration by its homography and a depth estimation problem showing the use of our differentiable warpers in a multi-scale fashion. Finally, we show an example making use of our differentiable local features implementations to solve a classical wide baseline stereo matching problem.

\subsection{Batch image processing}
\label{section:use_cases:imgproc}

In Section \ref{section:related_work} we reviewed existing libraries implementing classical image processing algorithms optimized for practical applications such noise reduction, image enhancement, and restoration. In this example we want show the utility of our framework for similar purposes. In addition, we include a benchmark comparing our framework to other existing vision libraries showing that even though \lib{} is not explicitly optimized for computer vision, similar results can be obtained in terms of performance.

As stated in section \ref{section:kornia:library_structure}, \lib{} provides implementations for low level processing e.g. color conversions, filtering and geometric image transformations that implicitly use native PyTorch operators such as 2D convolutions and simple matrix multiplications  all  optimized for CPU and GPU usage. Qualitative results of our image processing API are illustrated in figure \ref{fig:imgproc}. Our API can be combined with other PyTorch components allowing to run vision algorithms via parallel programming, or even sending composed functions to distributed environments. In Figure \ref{fig:imgproc:listing_benchmark}, we provide Python code highlighting the simplicity of our API and how, with very few lines of code, we can create a composed function to compute the Sobel edges~\cite{kanopoulos1988design} of a given batch of images transparent to the device or even send the composed function to a distributed set of devices in order to build applications at large-scale, or for just simply do the data augmentation in the GPU.

\textbf{Benchmark.} The scope of this library is to not provide explicitly optimized code for vision, but we want to show an experiment comparing the performance of our library with respect to other existing vision libraries e.g. OpenCV \cite{opencv}, PIL, skimage \cite{scikit-image} and scipy \cite{scikit-learn}, see figure \ref{fig:imgproc:listing_benchmark}. The purpose of this experiment is to not give a detailed benchmark between frameworks, but just to have an idea of how our implementations compares to libraries that are very well optimized for computer vision. The setup of the experiment assumes as input an RGB tensor of images with a fixed resolution of (256x256) varying the size of the batch. In this experiment, we compute Sobel edges 500 times measuring the median elapsed time between samples. The results show that for small batches, \lib's performance is similar to those obtained using  other libraries. It is worth noting that when we use a large batch size, the performance for our CPU implementation is the lowest, but when using the GPU we get the best timing performance. The machine used for this experiment was an Intel(R) Xeon(R) CPU E5-1620 v3 @ 3.50GHz and a Nvidia Geforce GTX 1080 Ti.

\begin{figure*}
    \setlength\tabcolsep{2.5pt}
    \begin{center}
        \begin{tabular}{c c c c c c c c}
        \textbf{Level 1} & \textbf{Level 2} & \textbf{Level 3} & \textbf{Level 4} & \textbf{Level 5} & \textbf{Level 6} \\
        \includegraphics[width=2.7cm]{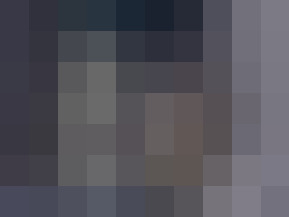} &
        \includegraphics[width=2.7cm]{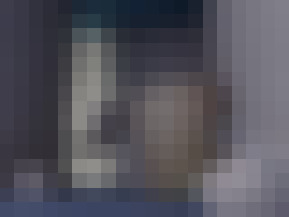} &
        \includegraphics[width=2.7cm]{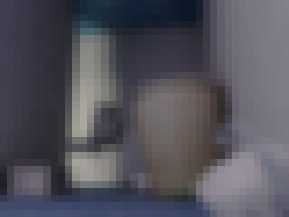} &
        \includegraphics[width=2.7cm]{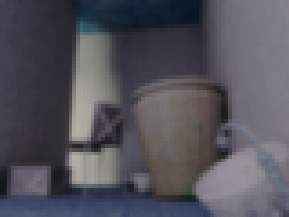} &
        \includegraphics[width=2.7cm]{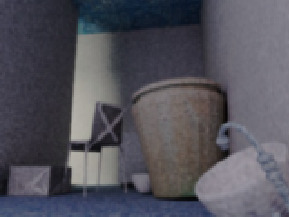} &
        \includegraphics[width=2.7cm]{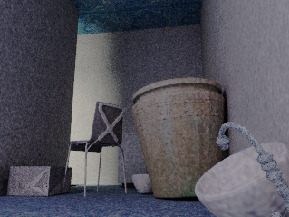} \\
        \includegraphics[width=2.7cm]{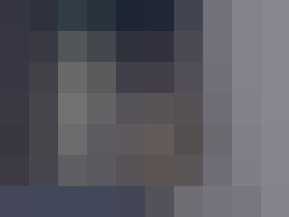} &
        \includegraphics[width=2.7cm]{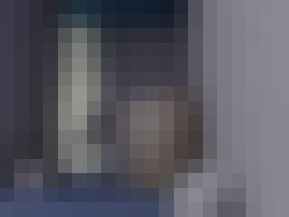} &
        \includegraphics[width=2.7cm]{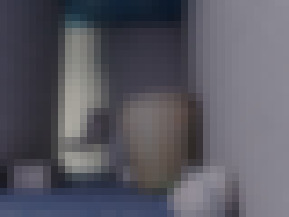} &
        \includegraphics[width=2.7cm]{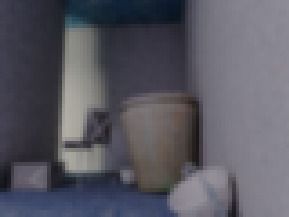} &
        \includegraphics[width=2.7cm]{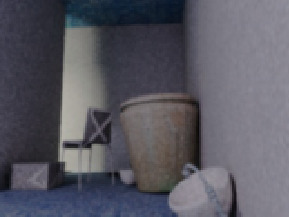} &
        \includegraphics[width=2.7cm]{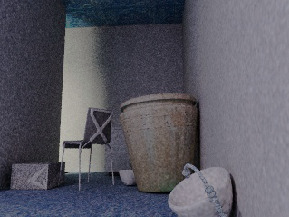} \\
        \includegraphics[width=2.7cm]{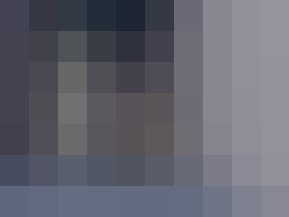} &
        \includegraphics[width=2.7cm]{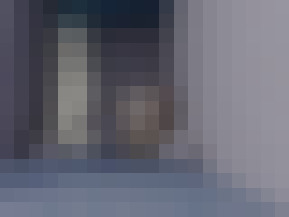} &
        \includegraphics[width=2.7cm]{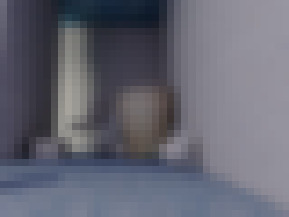} &
        \includegraphics[width=2.7cm]{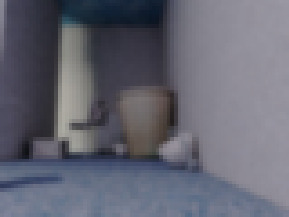} &
        \includegraphics[width=2.7cm]{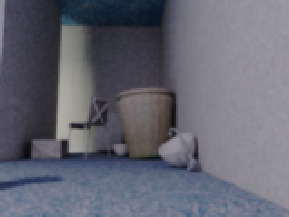} &
        \includegraphics[width=2.7cm]{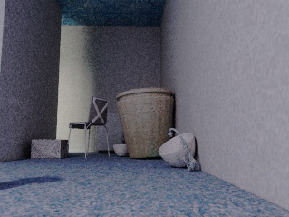} \\
        \includegraphics[width=2.7cm]{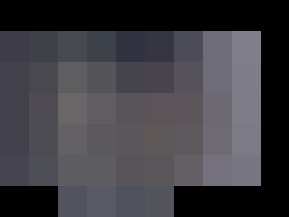} &
        \includegraphics[width=2.7cm]{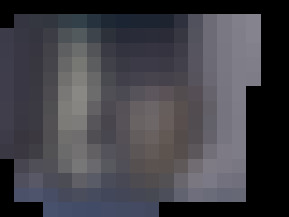} &
        \includegraphics[width=2.7cm]{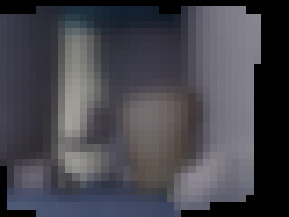} &
        \includegraphics[width=2.7cm]{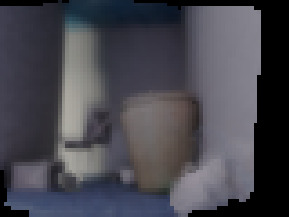} &
        \includegraphics[width=2.7cm]{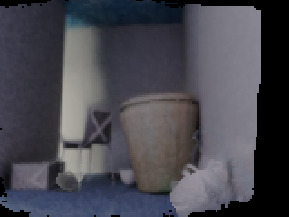} &
        \includegraphics[width=2.7cm]{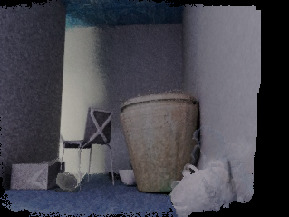} \\
        \includegraphics[width=2.7cm]{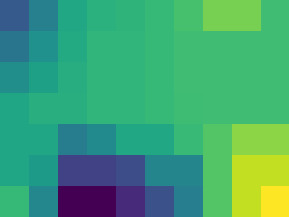} &
        \includegraphics[width=2.7cm]{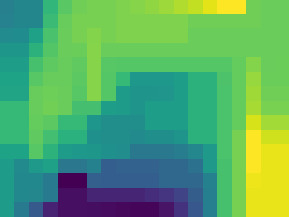} &
        \includegraphics[width=2.7cm]{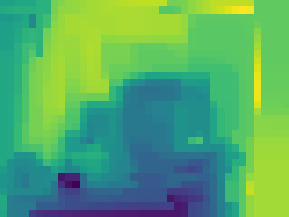} &
        \includegraphics[width=2.7cm]{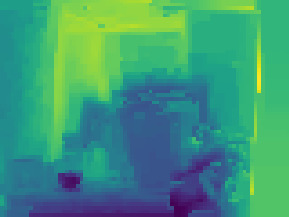} &
        \includegraphics[width=2.7cm]{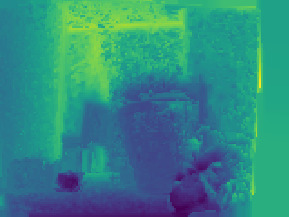} &
        \includegraphics[width=2.7cm]{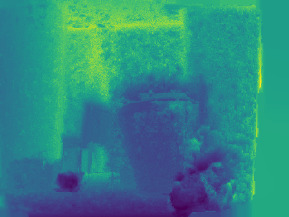} \\
        \includegraphics[width=2.7cm]{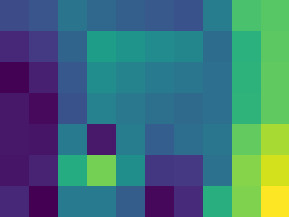} &
        \includegraphics[width=2.7cm]{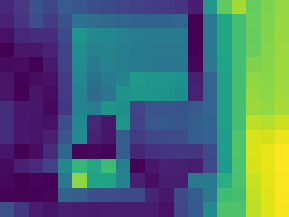} &
        \includegraphics[width=2.7cm]{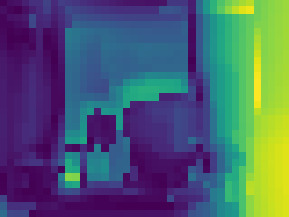} &
        \includegraphics[width=2.7cm]{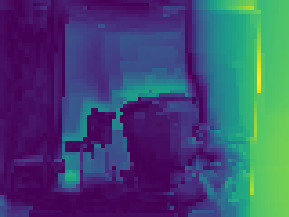} &
        \includegraphics[width=2.7cm]{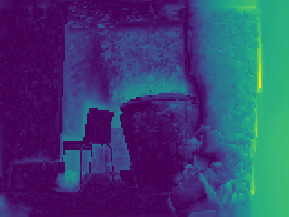} &
        \includegraphics[width=2.7cm]{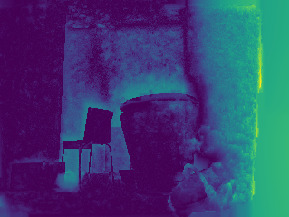} &
        \end{tabular}
    \end{center}
    \caption{Results of the depth estimation by gradient descent showing the depth map produced by the given set of calibrated camera images over different scales. Each column represents a level of a multi-resolution image pyramid. \textit{Row 1 to 3:} the source images, where the 2\textit{nd} row is the reference view; \textit{Row 3:} the images from row 1 and 3 warped to the reference camera given the depth at that particular scale level. \textit{Row 4 \& 5:} the estimated depth map and the error per pixel compared to the ground truth depth map in the reference camera. The data used for these experiments was extracted from SceneNet RGB-D dataset \protect\cite{McCormac:etal:ICCV2017}, containing photorealistic indoor image trajectories.}
    \label{fig:multiview:depth_estimation}
\end{figure*}

\subsection{Image registration by Gradient Descent}
\label{section:use_cases:homography_estimation}
In the following, we show the potential of the library for tasks requiring 2D planar geometry (for instance, marker-based camera pose estimation, spatial transformer networks, etc.). \lib{} provides a set of differentiable operators to perform geometric image transformations such as rotations, translations, scalings, shearings, as well as affine and homography transformation. At the core of the geometry module, we have implemented an operator  |kornia.HomographyWarper|, which warps by the homography a tensor in the reference frame A to a reference frame B that can be used to put in correspondence a set of images in a very efficient way.

\textbf{Implementation.} The task to solve is image registration using a multi-scale version of the Lucas-Kanade~\cite{BaM2004} strategy. Given a pair of images $I_a$ and $I_b$, it optimizes the parameters of the homography $H_a^b$ that minimizes the photometric error between $I_b$ and the transformation of $\hat{I_b}$ denoted as $\omega(I_a, H_a^b)$. Thanks to the Pytorch \textit{Autograd} engine this can be implemented without explicitly computing the derivatives of the loss function from equation \ref{eq:homography_loss}, resulting in a very compact and intuitive code.

\begin{equation}
\text{Loss} = \sum_{u,v}^{N} \| I_b - \omega(I_a, H_a^b) \|_1
\label{eq:homography_loss}
\end{equation}

The loss function is optimized at each level of a multi-resolution pyramid, from the lower to the upper resolution levels. Figure \ref{fig:homography_pyramid} shows the original images, warped images and the error per pixel with respect to the ground truth warp at each of the scale levels. We use the Adam~\cite{adam2015} optimizer with a learning rate of $1e-3$, iterating 200 times at each scale level. As a side note, pixel coordinates are normalized in the range of $[-1, 1]$, meaning that there is no need to re-scale the optimized parameters between pyramid levels. 

\label{section:use_cases:depth_estimation}
\subsection{Multi-View Depth Estimation by Gradient Descent}
In this example we have implemented a fully differential generic multi-view pipeline, using our framework, to allow for systematically using multi-view video data for machine learning research and applications. For this purpose we provide the |kornia.DepthWarper| operator that takes an arbitrary number of calibrated camera views and warps them to a reference camera frame given the depth in the reference frame.

Multi-view reconstruction is a well understood problem with a good geometric model \cite{Hartley_MVG}, and many approaches for matching and optimization \cite{dtam_Newcombe2011, patchmatch-stereo, Sevilla-LaraSJB16}, and some recent promising deep learning approaches \cite{Luo2016}. We have found current machine learning approaches \cite{FischerDIHHGSCB15, IlgMSKDB16} to be limiting, in that they have not been generalized to arbitrary numbers of views (spatial or temporal); and available datasets \cite{Butler2012, Geiger2012CVPR} are only stereo and low resolution. Many of the machine learning approaches assume that there is high quality ground truth depth provided as commonly available datasets, which limits
their application to new datasets or new camera configurations. Classical approaches such as planesweep, patch match or DTAM~\cite{Newcombe:2011:DDT:2355573.2356447} have not been implemented with deep learning in mind, and do not fit easily into existing deep learning frameworks.

\textbf{Implementation.} We start with a simple formulation that allows us to solve for depth images using gradient descent with a variety of losses based on state of the art classical approaches (photometric, depth consistency, depth piece-wise smoothness, and multi-scale pyramids).

The multi-view reconstruction pipeline receives as input a set of views, with RGB images and calibrated intrinsic camera models, $K_{i}$, and pose estimates $T_{\text{ref}}^{i}$, and then solves for the depth image, $\boldsymbol{d}_{\text{ref}}$, for a reference view. Since we assume a calibrated setup, the depth value of a given pixel $\boldsymbol{u}_{\text{ref}} = [u_{\text{ref}},v_{\text{ref}}]$ in the reference view, $\boldsymbol{d}_{\text{ref}}$, can be used to compute the corresponding pixel location, $\boldsymbol{u}_i = [u_{i},v_{i}]$ in any of the other views through simple projective geometry $H_{\text{ref}}^{i} = K_i \cdot T_{\text{ref}}^{i} \cdot K_{\text{ref}}^{-1}$. Given this, we can warp views onto each other parameterized by depth and camera calibration using a differentiable bilinear sampling as proposed in~\cite{stn_NIPS2015}, $\tilde{I_{\text{ref}}} = \omega(I_i, H_{\text{ref}}^{i}, \boldsymbol{d}_{\text{ref}})$.

Similar to \cite{dtam_Newcombe2011, monodepth17, superdepth18}, depth is solved for by minimizing a photometric error between the views warped to the reference view, (equation ~\ref{eq:multiview:loss photometric1} and ~\ref{eq:multiview:loss photometric2}). We compute an additional loss to encourage disparities to be locally smooth with a penalty on the disparity gradients weighted by image gradients as seen in equation~\ref{eq:multiview:loss smoothness}. Finally, losses are combined with a weighted sum (see in equation \ref{eq:multiview:loss total}). These losses are easily modified or extended, depending on how well the assumptions about these losses fit the data, e.g. it is naive and assumes photometric consistency which is only true for small view displacements.
\begin{align}
	\label{eq:multiview:loss photometric1}
	L_{\text{photo1}} &= \frac{1}{n} \sum\limits^{n} \dfrac{1 - \text{SSIM}(I_{\text{ref}}, \tilde{I_{\text{ref}}})}{2} \\
	\label{eq:multiview:loss photometric2}
	L_{\text{photo2}} &= \frac{1}{n} \sum\limits^{n} \lvert I_{\text{ref}} - \tilde{I_{\text{ref}}} \rvert \\
	\label{eq:multiview:loss smoothness}	
	L_{\text{smooth}} &= \frac{1}{n} \sum\limits^{n} \lvert \partial_{x} \, d \rvert e^{- \| \partial_{x} I_{i} \|} + \lvert \partial_{y} \, d \rvert e^{- \| \partial_{y} I_{i} \|} \\
	\label{eq:multiview:loss total}
	L_{\text{total}} &= \alpha L_{\text{photo1}} + (1 - \alpha) L_{\text{photo2}} + \lambda L_{\text{smooth}}
\end{align}
Figure~\ref{fig:multiview:depth_estimation} shows partial results obtained by the depth algorithm implemented using \lib. The algorithm receives as input 3 calibrated views with RGB images (320x240). We used Stochastic Gradient Descent (SGD) with momentum and compute the depth at 7 different scales by blurring the image and down-sampling the resolution by a factor of 2 from the previous size. To compute the loss, we up-sample again to the original size using bilinear interpolation.  The refinement at each level was done for 500 iterations starting from the lowest resolution and going up. The initial values for depth were obtained by a random uniform sampling in a range between 0 and 1.
\subsection{Targeted adversarial attack on two view matching with SIFT}
\label{section:use_cases:adversarial_matching}
\begin{figure}[t]
    \begin{center}
        \includegraphics[width=0.45\linewidth]{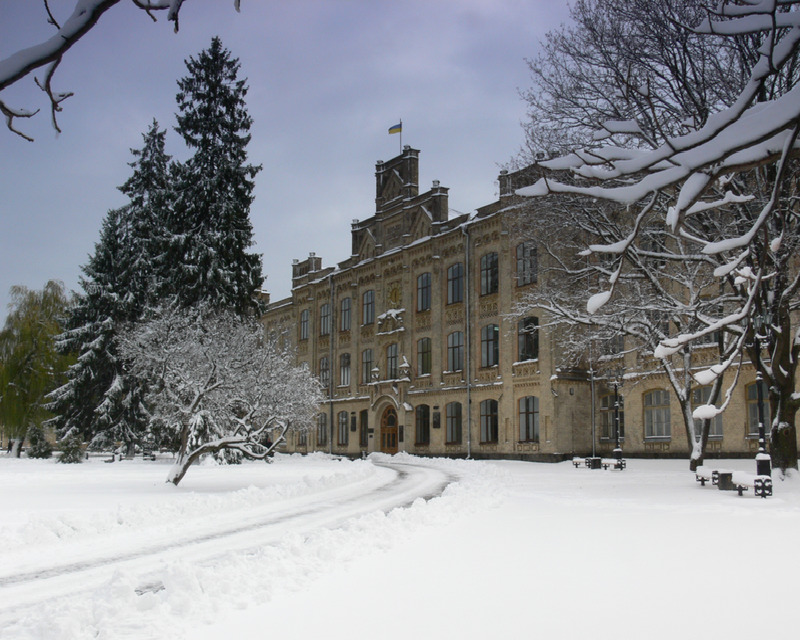} 
        \includegraphics[width=0.45\linewidth]{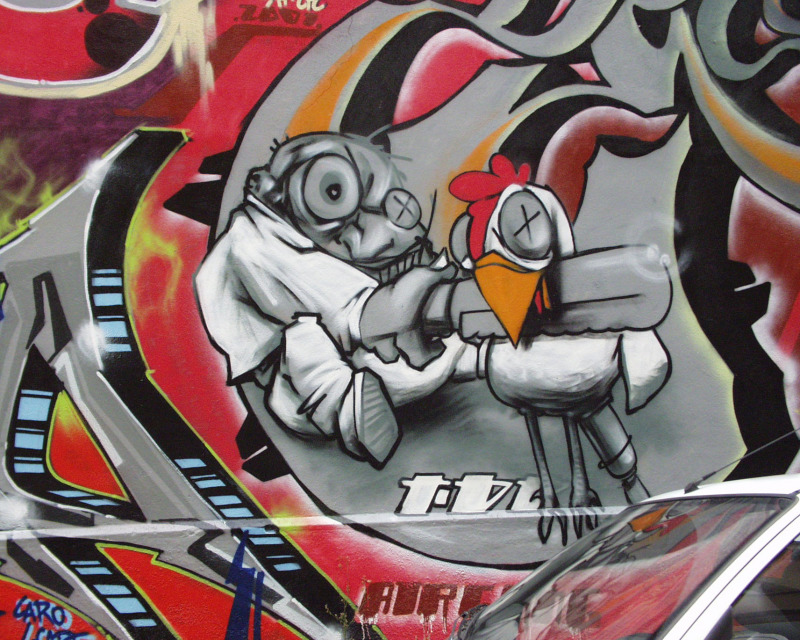} \\
        \includegraphics[width=0.45\linewidth]{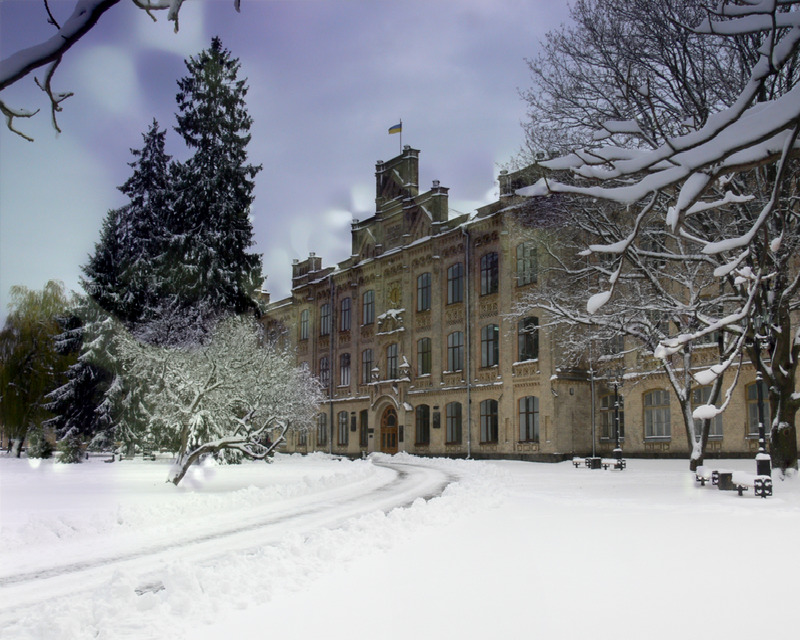} 
        \includegraphics[width=0.45\linewidth]{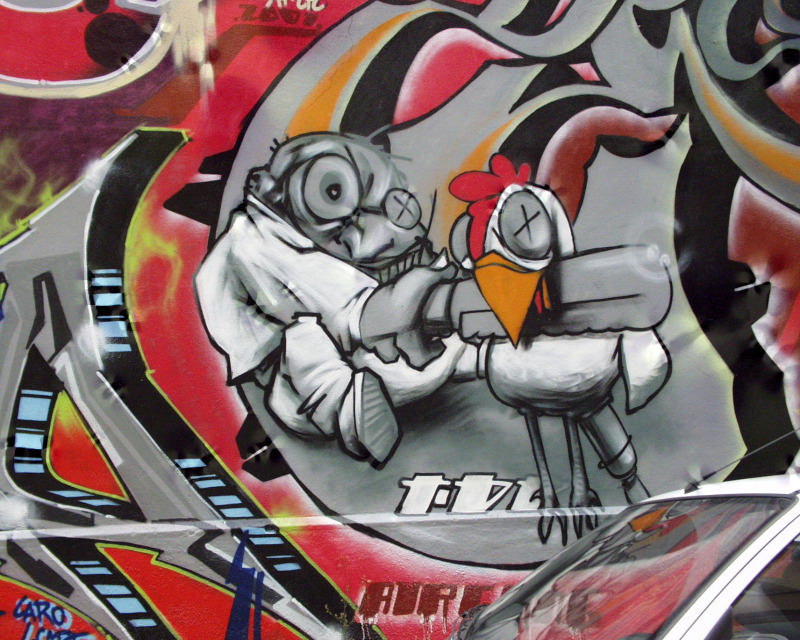} \\
        \includegraphics[width=0.45\linewidth]{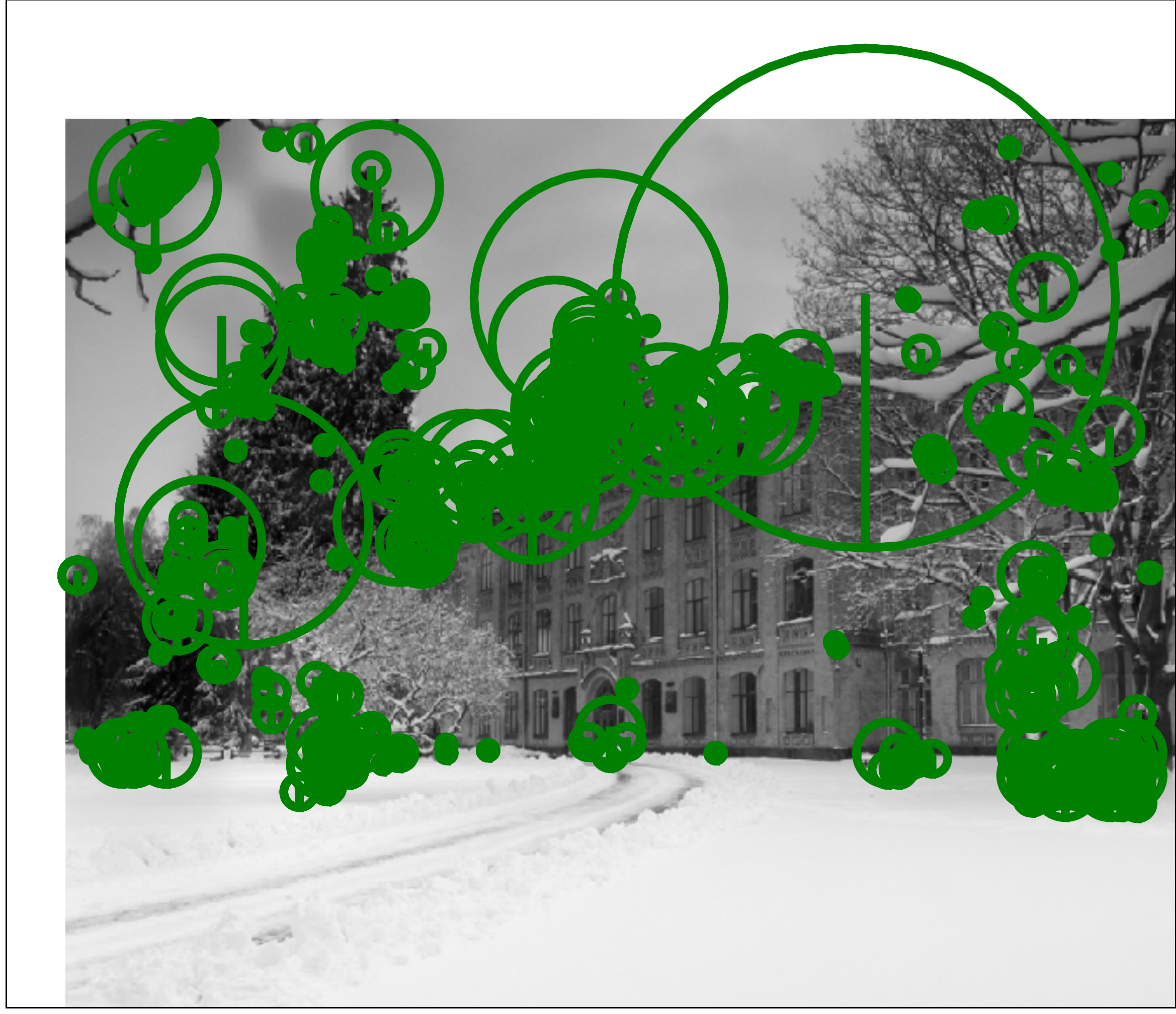} 
        \includegraphics[width=0.45\linewidth]{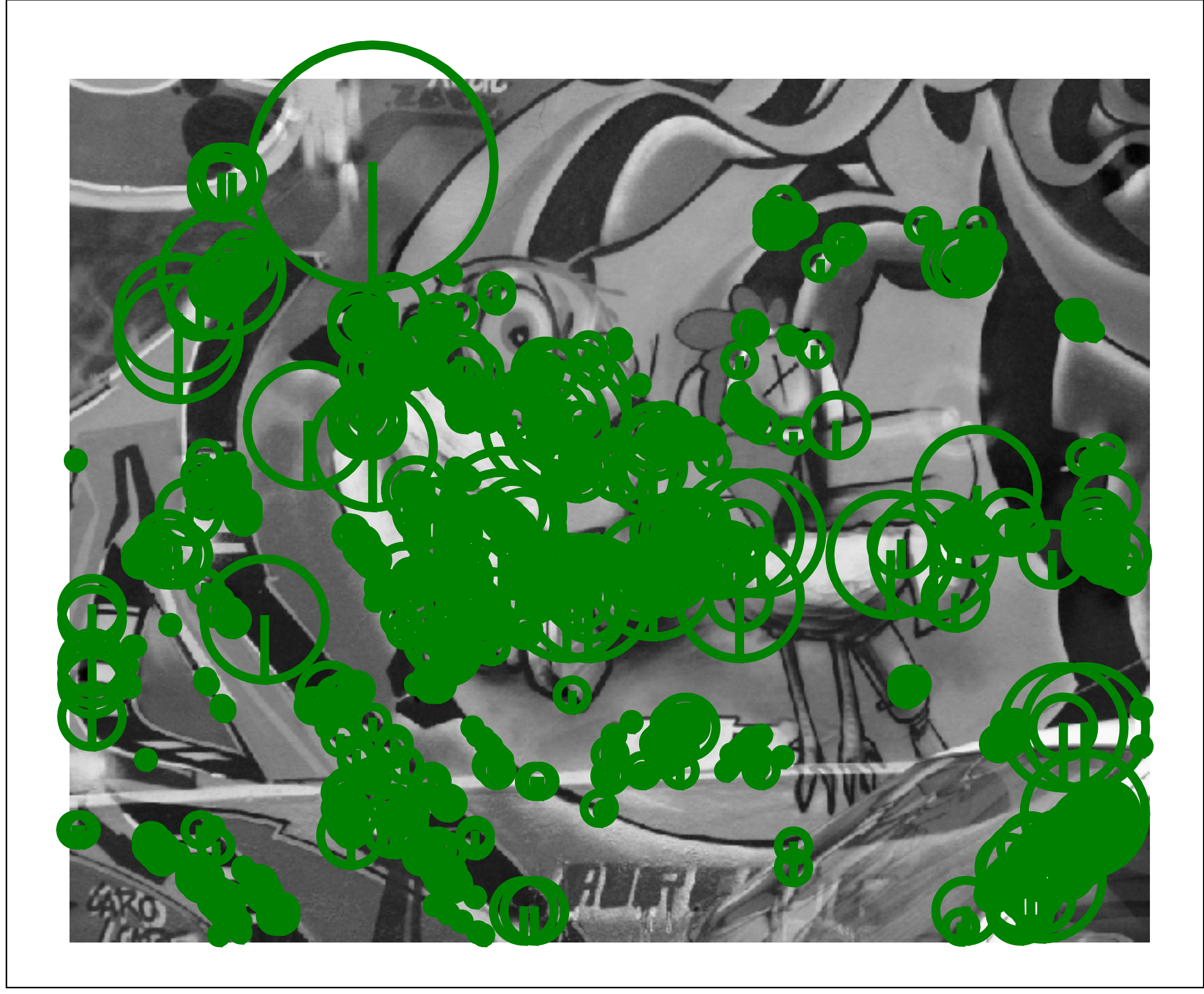} \\
        \includegraphics[width=0.45\linewidth]{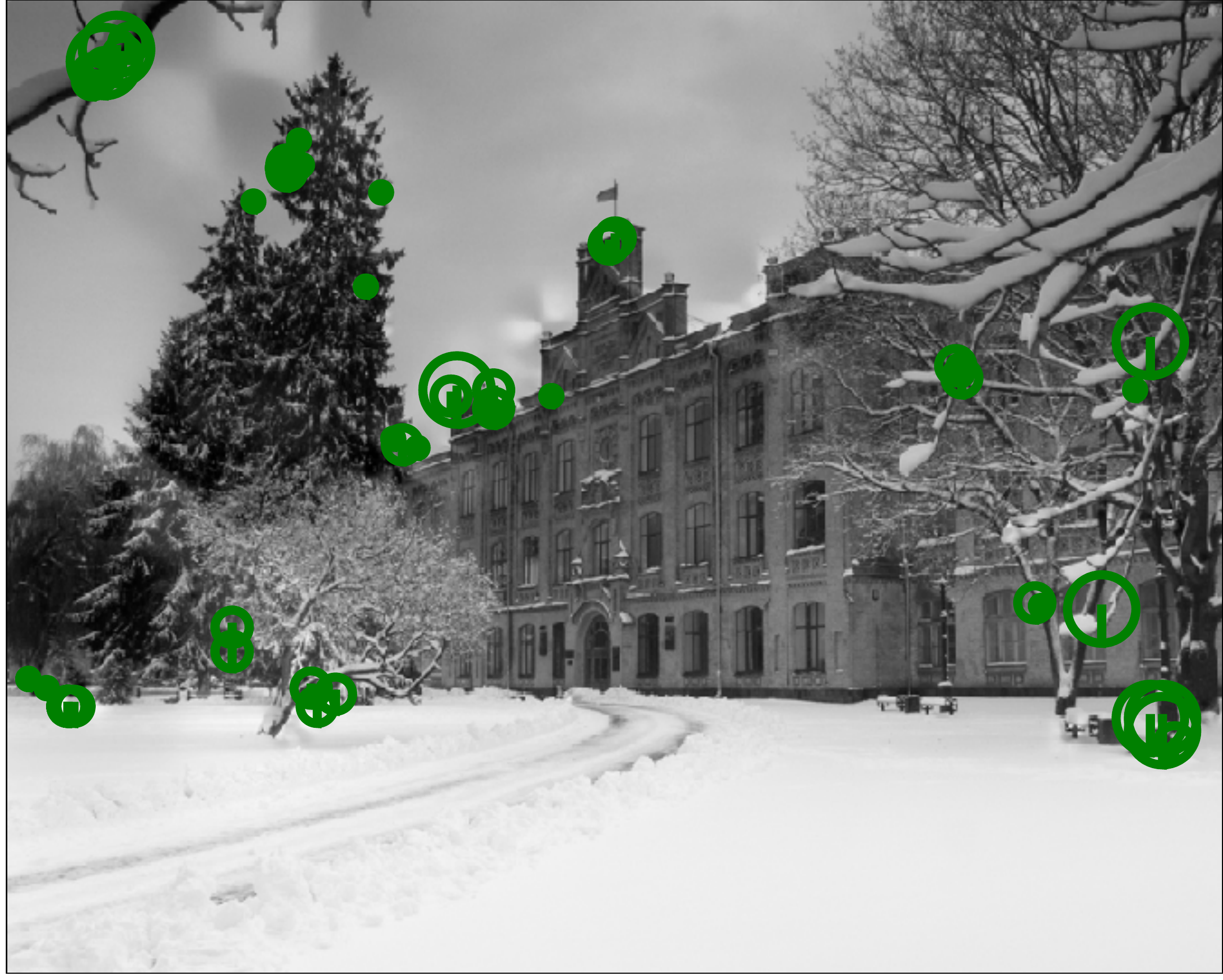} 
        \includegraphics[width=0.45\linewidth]{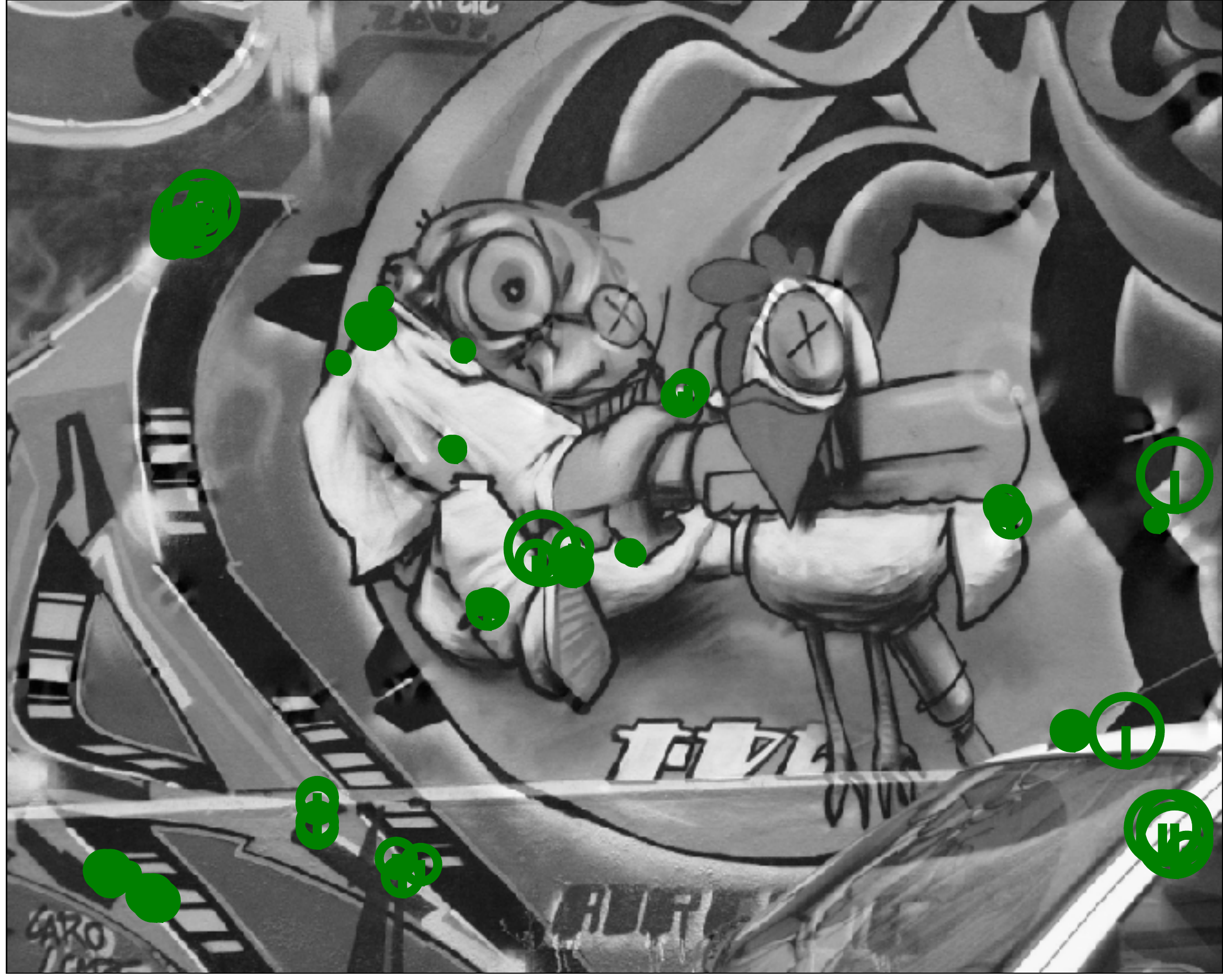} \\
        
    \end{center}
    \caption{Targeted adversarial attack on image matching. From top to bottom: original images, which do not match; images, optimized by gradient descent to have local features that match; the result of the attack: matching features (Hessian detector + SIFT descriptor). Matching features, which survived RANSAC geometric verification}
    \label{fig:wbs}
\end{figure}
In the following example we show how to implement fully differential wide baseline stereo matching with local feature detectors and descriptors using |kornia.features|. We demonstrate the differentiability by making a targeted adversarial attack on the wide baseline matching pipeline.

\textbf{Local feature detectors and descriptors}
Local features are the workhorses of 3d reconstruction~\cite{schonberger2016structure, torii2018structure}, visual localization~\cite{sarlin2019coarse} and image retrieval \cite{shen2018matchable}. Although learning-based methods now seems to dominate~\cite{LocaFeaturesReview2018}, recent benchmark top-performers still use Difference-of-Gaussians aka SIFT detector~\cite{CVPRW2019}. SIFT descriptor is still one of the best for 3d reconstruction~\cite{ColmapBenchmark2017} tasks. Thus, we believe that community would benefit from having GPU-accelerated and differentiable version of the classical tools. 

\textbf{Adversarial attacks.}
Adversarial attacks is an area of research which recently gained popularity after the seminal work of Szegedy et al.~\cite{AdvAttack2014} showing that small perturbations in the input image can switch the neural network prediction outcome. There are series of works showing that CNN-based solution of classification~\cite{NIPS2018Adv}, segmentation~\cite{arnab_cvpr_2018}, object detection~\cite{ObjDetAdv2018}, and image retrieval~\cite{AdvRetrieval2019} tasks are all prone to such attacks. Yet, the authors do not know of any paper devoted to adversarial attacks on local features-based image matching. 
Most of attack methods are "white-box"~\cite{NIPS2018Adv}, which means they require access to the model gradients w.r.t the input. This makes them an excellent choice for a |kornia.features| differentiability demonstration.

\textbf{Implementation.}
The two view matching task is posed in a following way~\cite{Pritchett1998}: given two images $I_{a}$ and $I_{b}$ depicting the same scene, find the correspondences between pixels in images. If $I_{a}$ and $I_{b}$ do not depict the same scene, no correspondences should be returned. This is typically solved by detecting local features, describing the local patches with descriptor and then matching by minimum descriptor distance with some filtering. \lib{} has all these parts implemented. 

We consider the following adversarial attack: given the non-matching image pair $I_{a}$, $I_{b}$, and the desired homography $H_a^b$, modify images so that the correspondence finding algorithm will output a non-negligible number of matches consistent with the homography $H_a^b$. This means that both local detectors should fire in specific locations and the local patches around that location should be matchable by given function: 

\begin{align}
	\label{eq:adversarial:loss total}
	L_{\text{total}} &= L_{\text{loc}} + \alpha L_{\text{desc}} + \beta L_{\text{reg}}\\
	\label{eq:adversarial:loss loc}
	L_{\text{loc}} &= \frac{1}{n} \sum\limits^{n} (p_1 - H p_2)^2 \\
   \label{eq:adversarial:loss desc}
	L_{\text{desc}} &= \frac{1}{n} \sum\limits^{n} (1 + d(D_1, D_2) - d(D_1, D_{2neg})) \\
	\label{eq:adversarial:loss reg}
	L_{\text{reg}} &= \frac{1}{n} \sum\limits^{n} (I - I_{\text{init}})^2 
\end{align}
where $p_1$ is keypoint detected in $I_a$, $p_2$ is closest reprojected by the $H_a^b$ keypoint detected in image $I_b$, $\sigma_1$ and $\sigma_2$ are their scales, $D_1$ and $D_2$ -- their descriptors, $D_{2neg}$ - hard negative in batch, $d(\cdot, \cdot)$ -- L2 distance, and $I_{\text{init}}$ is original unmodified version of $I_a$ and $I_b$.

The detector used in the example is the Hessian blob detector~\cite{Hessian78}; the descriptor is the SIFT~\cite{Lowe2004}. We keep the top-2500 keypoints and use the Adam~\cite{adam2015} optimizer with a learning rate of 0.003. 
Figure~\ref{fig:wbs} shows the original images, optimized images and optimized images with matching features visualized.  The perturbations are not quite imperceptible, but that it is not the goal of the current example.  

\section{Conclusions}
We have introduced \lib, a library for computer vision in PyTorch that implements traditional vision algorithms in a differentiable fashion making use of the hardware acceleration to improve the performance. We demonstrated how by using our library, classical vision problems such as image registration by homography, depth estimation, or local features matching can be very easily solved with a high performance similar to existing libraries. By leveraging this project, we believe that classical computer vision libraries can take a different role within the deep learning environments as components of layers of the networks as well as pre- and post-processing of the results. In the future, we expect researchers and companies increase the number of such contributions. At the time of submission, \lib{} has 660 github stars and 60 forks
%
\section{Acknowledgement}
We would like to acknowledge Arraiy, Inc. to sponsor the initial stage of the project. The folks from the OSVF/OpenCV.org and the PyTorch open-source community for helpful contributions and feedback. The work of Edgar Riba and Daniel Ponsa has been partially supported by the Spanish Government under
Project TIN2017-89723-P. Dmytro Mishkin is supported by CTU student grant SGS17/185/OHK3/3T/13 and by the Austrian Ministry for Transport, Innovation and Technology, the Federal Ministry of Science, Research and Economy, and the Province of Upper Austria in the frame of the COMET center SCCH.

{\small
\bibliographystyle{ieee}
\bibliography{egbib}
}

\end{document}